\def\year{2021}\relax
\documentclass[letterpaper]{article} 
\usepackage{aaai21}  
\usepackage{times}  
\usepackage{helvet} 
\usepackage{courier}  
\usepackage[hyphens]{url}  
\usepackage{graphicx} 
\usepackage{natbib}  
\usepackage{caption} 
\usepackage{algorithm,algorithmicx}
\usepackage[noend]{algpseudocode}
\usepackage{amsmath}
\usepackage{bbm}
\usepackage{bm}
\usepackage{amsfonts}
\usepackage{booktabs}
\usepackage{amsthm,amsmath,amssymb}
\usepackage{multirow}
\usepackage{makecell} 
\usepackage{mathrsfs}
\usepackage{subfigure}
\title{Contrastive and Generative Graph Convolutional Networks for Graph-based Semi-Supervised Learning}
\author{Sheng Wan\textsuperscript{\rm 1}, Shirui Pan\textsuperscript{\rm 2}, Jian Yang\textsuperscript{\rm 1}, Chen Gong\textsuperscript{\rm 1}\thanks{Corresponding author}}
\affiliations{

    \textsuperscript{\rm1}School of Computer Science and Engineering, Nanjing University of Science and Technology, Nanjing, China\\
    \textsuperscript{\rm2}Faculty of IT, Monash University, Australia

}

\begin{document}

\maketitle

\begin{abstract}
Graph-based \textbf{S}emi-\textbf{S}upervised \textbf{L}earning (SSL) aims to transfer the labels of a handful of labeled data to the remaining massive unlabeled data via a graph. As one of the most popular graph-based SSL approaches, the recently proposed \textbf{G}raph \textbf{C}onvolutional \textbf{N}etworks (GCNs) have gained remarkable progress by combining the sound expressiveness of neural networks with graph structure. Nevertheless, the existing graph-based methods do not directly address the core problem of SSL, \emph{i.e.}, the shortage of supervision, and thus their performances are still very limited. To accommodate this issue, a novel GCN-based SSL algorithm is presented in this paper to enrich the supervision signals by utilizing both data similarities and graph structure. Firstly, by designing a semi-supervised contrastive loss, improved node representations can be generated via maximizing the agreement between different views of the same data or the data from the same class. Therefore, the rich unlabeled data and the scarce yet valuable labeled data can jointly provide abundant supervision information for learning discriminative node representations, which helps improve the subsequent classification result. Secondly, the underlying determinative relationship between the data features and input graph topology is extracted as supplementary supervision signals for SSL via using a graph generative loss related to the input features. Intensive experimental results on a variety of real-world datasets firmly verify the effectiveness of our algorithm compared with other state-of-the-art methods.
\end{abstract}

\section{Introduction}

\textbf{S}emi-\textbf{S}upervised \textbf{L}earning (SSL) focuses on utilizing small amounts of labeled data as well as relatively large amounts of unlabeled data for model training \cite{zhu2005semi}. Over the past few decades, SSL has attracted increasing research interests and a variety of approaches have been developed \cite{zhu2003semi, joachims1999transductive}, which usually employ cooperative training \cite{blum1998combining}, support vector machines \cite{bennett1999semi, li2010s4vm}, consistency regularizers \cite{tarvainen2017mean, laine2016temporal, berthelot2019mixmatch}, and graph-based methods \cite{zhu2003semi, gong2015deformed, belkin2006manifold, Kipf2016Semi, ma2019flexible}. Among them, the graph-based SSL algorithms have gained much attention due to its ease of implementation, solid mathematical foundation, and satisfactory performance.

In a graph-based SSL algorithm, all labeled and unlabeled data are represented by graph nodes and their relationships are depicted by graph edges. Then the problem is to transfer the labels of a handful of labeled nodes (\emph{i.e.}, labeled examples) to the remaining massive unlabeled nodes (\emph{i.e.}, unlabeled examples) such that the unlabeled examples can be accurately classified. A popular method is to use graph Laplacian regularization to enforce the similar examples in the feature space to obtain similar label assignments, such as \cite{belkin2006manifold, gong2015deformed, buhler2009spectral}. Recently, research attention has been shifted to the learning of proper network embedding to facilitate the label determination \cite{Kipf2016Semi, defferrard2016convolutional, zhou2019graph, velivckovic2018graph, hu2019hierarchical, yan2018spatial}, where \textbf{G}raph \textbf{C}onvolutional \textbf{N}etworks (GCNs) have been demonstrated to outperform traditional graph-based models due to its impressive representation ability \cite{wu2020comprehensive}. Concretely, GCNs generalize \textbf{C}onvolutional \textbf{N}eural \textbf{N}etworks (CNNs) \cite{lecun1995convolutional} to graph-structured data based on the spectral theory, and thus can reconcile the expressive power of graphs in modeling the relationships among data points for representation learning.

Although graph-based SSL methods have achieved noticeable progress in recent years, they do not directly tackle the core problem of SSL, namely the shortage of supervision. One should note that the number of labeled data in SSL problems is usually very limited, which poses a great difficulty for stable network training and thus will probably degrade the  performance of GCNs. To accommodate this issue, in this paper, we aim at sufficiently extracting the supervision information carried by the available data themselves for network training, and develop an effective transductive SSL algorithm via using GCNs. That is to say, the goal is to accurately classify the observed unlabeled graph nodes \cite{zhu2005semi, gong2016label}.

Our proposed method is designed to enrich the supervision signals from two aspects, namely data similarities and graph structure. Firstly, considering that the similarities of data points in the feature space provide the natural supervision signals, we propose to use the recently developed contrastive learning \cite{he2020momentum, chen2020simple} to fully explore such information. Contrastive learning is an active field of self-supervised learning \cite{doersch2015unsupervised, gidaris2018unsupervised}, which is able to generate data representations by learning to encode the similarities or dissimilarities among a set of unlabeled examples \cite{hjelm2018learning}. The intuition behind is that the rich unlabeled data themselves can be used as supervision signals to help guide the model training. However, unlike the typical unsupervised contrastive learning methods \cite{hassani2020contrastive, velickovic2019deep}, SSL problem also contains scarce yet valuable labeled data, so here we design a new semi-supervised contrastive loss, which additionally incorporates class information to improve the contrastive representation learning for node classification tasks. Specifically, we obtain the node representations generated from global and local views respectively, and then employ the semi-supervised contrastive loss to maximizing the agreement between the representations learned from these two views. Secondly, considering that the graph topology itself contains precious information which can be leveraged as supplementary supervision signals for SSL, we utilize a generative term to explicitly model the relationship between graph and node representations. As a result, the originally limited supervision information of labeled data can be further expanded by exploring the knowledge from both data similarities and graph structure, and thus leading to the improved data representations and classification results. Therefore, we term the proposed method as `\textbf{C}ontrastive \textbf{G}CNs with \textbf{G}raph \textbf{G}eneration' (CG$^3$). In experiments, we demonstrate the contributions of the supervision clues from utilizing contrastive learning and graph structure, and the superiority of our proposed CG$^3$ to other state-of-the-art graph-based SSL methods has also been verified.

\section{Related Work}

In this section, we review some representative works on graph-based SSL and contrastive learning, as they are related to this article.

\subsection{Graph-based Semi-Supervised Learning}

Graph-based SSL has been a popular research area in the past two decades. Early graph-based methods are based on the simple assumption that nearby nodes are likely to have the same label. This purpose is usually achieved by the low-dimensional embeddings with Laplacian eigen-maps \cite{belkin2004semi, belkin2006manifold}, spectral kernels \cite{zhang2006analysis}, Markov random walks \cite{szummer2002partially, zhou2004learning, gong2014fick}, \emph{etc}. Another line is based on graph partition, where the cuts should agree with the class information and are placed in low-density regions \cite{zhu2002learning, speriosu2011twitter}. In addition, to further improve the learning performance, various techniques are proposed to jointly model the data features and graph structure, such as deep semi-supervised embedding \cite{weston2012deep} and Planetoid \cite{yang2016revisiting} which regularize a supervised classifier with a Laplacian regularizer or an embedding-based regularizer. Recently, a set of graph-based SSL approaches have been proposed to improve the performance of the above-mentioned techniques, including \cite{calder2020poisson, gong2019multi, calder2019properly}.

Subsequently, inspired by the success of CNNs on grid-structured data, various types of graph convolutional neural networks have been proposed to extend CNNs to graph-structured data and have demonstrated impressive results in SSL \cite{dehmamy2019understanding, zhang2019bayesian, xu2019graph}. Generally, graph convolution can be attributed to the spatial methods directly working on node features and the spectral methods based on convolutions on nodes. In spatial methods, the convolution is defined as a weighted average function over the neighbors of each node which characterizes the impact exerting to the target node from its neighboring nodes, such as GraphSAGE \cite{hamilton2017inductive}, graph attention network (GAT) \cite{velivckovic2018graph}, and the Gaussian induced convolution model \cite{jiang2019gaussian}. Different from the spatial methods, spectral graph convolution is usually based on eigen-decomposition, where the locality of graph convolution is considered by spectral analysis \cite{jiang2019semi}. Concretely, a general graph convolution framework based on graph Laplacian is first proposed in \cite{bruna2014spectral}. Afterwards, ChebyNet \cite{defferrard2016convolutional} optimized the method by using Chebyshev polynomial approximation to realize eigenvalue decomposition. Besides, \cite{Kipf2016Semi} proposed GCN via using a localized first-order approximation to ChebyNet, which brings about more efficient filtering operations than spectral CNNs. Despite the noticeable achievements of these graph-based semi-supervised methods in recent years, the main concern in SSL, \emph{i.e.}, the shortage of supervision information, has not been directly addressed.

\subsection{Contrastive Learning}

Contrastive learning is a class of self-supervised approaches which trains an encoder to be contrastive between the representations that depict statistical dependencies of interest and those that do not \cite{velickovic2019deep, chen2020simple, tschannen2019mutual}. In computer vision, a large collection of works \cite{hadsell2006dimensionality, he2020momentum, tian2019contrastive} learn self-supervised representations of images via minimizing the distance between two views of the same image. Analogously, the concept of contrastive learning has also become the central to some popular word-embedding methods, such as word2vec model \cite{mikolov2013distributed} which utilizes co-occurring words and negative sampling to learn the word embeddings. 

Recently, contrastive methods can be found in several graph representation learning algorithms. For instance, \textbf{D}eep \textbf{G}raph \textbf{I}nfomax (DGI) \cite{velickovic2019deep} extends deep Infomax \cite{hjelm2018learning} via learning node representations through contrasting node and graph encodings. Besides, \cite{hassani2020contrastive} learns node-level and graph-level representations by contrasting the different structures of a graph. Apart from this work, a novel framework for unsupervised graph representation learning is proposed in \cite{zhu2020deep} by maximizing the agreement of node representations between two graph views. Although contrastive learning can use the data themselves to provide the supervision information for representation learning, they are not directly applicable to SSL as they fail to incorporate the labeled data which are scarce yet valuable in SSL. In this paper, we devise a semi-supervised contrastive loss function to exploit the supervision signals contained in both the labeled and unlabeled data, which can help learn the discriminative representations for accurate node classification.

\begin{figure*}[t]
	\centering
	\includegraphics[width=0.9\textwidth]{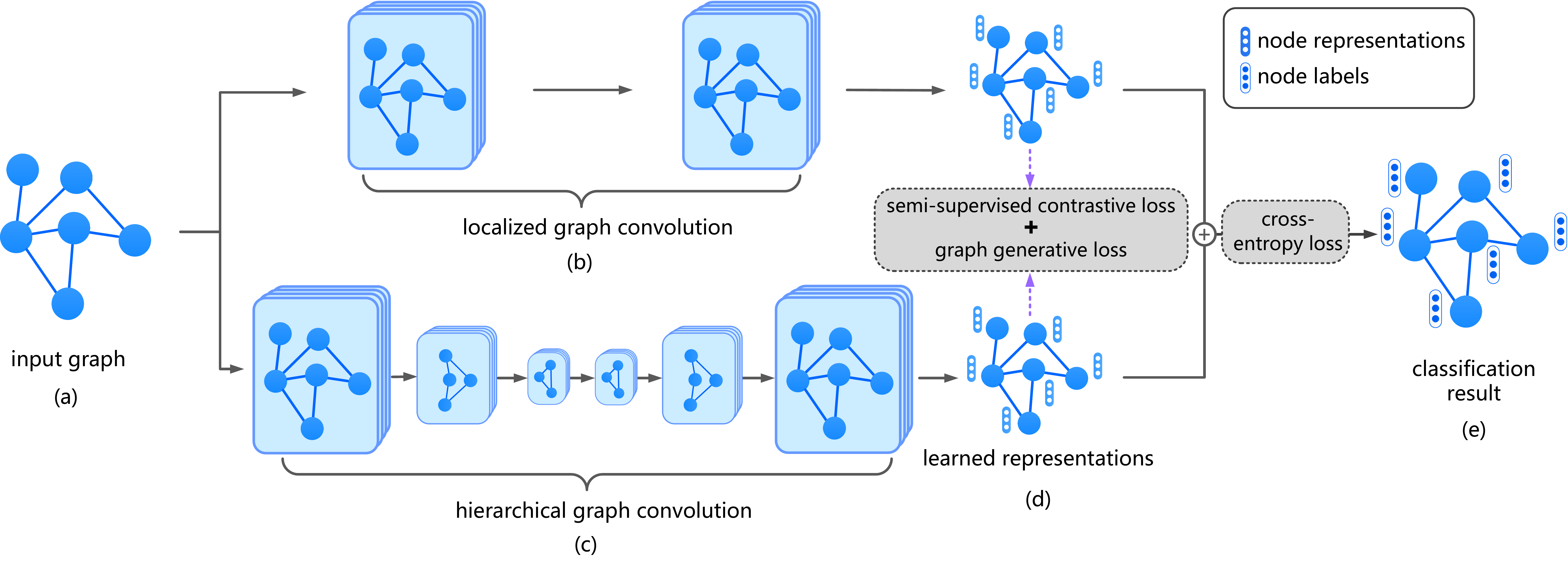} 
	\caption{The framework of our approach. In (a), the original graph is adopted as the input of (b) the localized GCNs and (c) the hierarchical GCNs, respectively, where (c) is utilized to capture the global information and serves as the augmented view of (b). In (d), the node representations are generated from (b) and (c), and then constitute the contrastive loss and graph generative loss collaboratively, in order to provide additional supervision signals to improve the representation learning process. In (e), the classification result is acquired via integrating the outputs of (b) and (c), where the cross-entropy loss is used to penalize the difference between the model prediction and the given labels of the initially labeled nodes.}
	\label{flowchart}
\end{figure*}

\section{Problem Description}

We start by formally introducing the problem of graph-based SSL. Suppose we have a set of $n=l+u$ examples $\Psi = \{\mathbf{x}_1, \cdots, \mathbf{x}_l, \mathbf{x}_{l+1}, \cdots, \mathbf{x}_n\}$, where the first $l$ examples constitute the labeled set with the labels $\{y_i\}_{i=1}^{l}$ and the remaining $u$ examples form the unlabeled set with typically $l \ll u$. We denote $\mathbf{X}\in\mathbb{R}^{n\times d}$ as the feature matrix with the $i$-th row formed by the feature vector $\mathbf{x}_i$ of the $i$-th example, and $\mathbf{Y}\in\mathbb{R}^{n\times c}$ as the label matrix with its $(i,j)$-th element $\mathbf{Y}_{ij}=1$ if $\mathbf{x}_i$ belongs to the $j$-th class and $\mathbf{Y}_{ij}=0$ otherwise. Here $d$ is the feature dimension and $c$ is the number of classes. The dataset $\Phi$ is represented by a graph $\mathcal{G}=\left\langle {\mathcal{V},\mathcal{E} } \right\rangle $, where $\mathcal{V}$ is the node set containing all examples and $\mathcal{E}$ is the edge set modeling the similarity among the nodes/examples. The adjacency matrix of $\mathcal{G}$ is denoted as $\mathbf{A}$ with $\mathbf{A}_{ij}=1$ if there exists an edge between $\mathbf{x}_i$ and $\mathbf{x}_j$ and $\mathbf{A}_{ij}=0$ otherwise. In this paper, we target transductive graph-based SSL which aims to find the labels $y_{l+1}, y_{l+2}, \cdots, y_n$ of the unlabeled examples $\mathbf{x}_{l+1}, \mathbf{x}_{l+2}, \cdots, \mathbf{x}_n$ based on $\Psi$.

\section{Method}

This section details our proposed CG$^3$ model (see Figure~\ref{flowchart}). Specifically, we illustrate the critical components of CG$^3$ by explaining the multi-view establishment for graph convolutions, presenting the semi-supervised contrastive learning, elaborating the graph generative loss, and describing the overall training procedure.

\subsection{Multi-View Establishment for Graph Convolutions}

In our CG$^3$ method, we need to firstly build two different views for the subsequent graph contrastive learning. Note that in self-supervised visual representation learning tasks, contrasting congruent and incongruent views of images enables the algorithms to learn expressive representations \cite{tian2019contrastive, he2020momentum}. However, unlike the regular grid-like image data where different views can be simply generated by standard augmentation techniques such as cropping, rotation, or color distortion, the view augmentation on irregular graph data is not trivial, as graph nodes and edges do not contain visually semantic contents as in the image \cite{velickovic2019deep}. Although edge removing or adding is a simple way to generate a related graph, it might damage the original graph topology, and thus degrading the representation results of graph convolutions. Instead of directly changing the graph structure, we employ two types of graph convolutions to generate node representations from two different views revealing the local and global cues. In this means, the representations generated by different views complement to each other and thus enriching the final representation results. Specifically, by performing contrastive learning between the obtained representations from two views, the rich global and local information can be encoded simultaneously. This process will be detailed as follows.

To obtain node representations from the local view, we actually have many choices of network architectures, such as the commonly-used GCN \cite{Kipf2016Semi} and GAT \cite{velivckovic2018graph} which produce node representations by aggregating neighborhood information. For simplicity, we adopt the GCN model \cite{Kipf2016Semi} as our backbone in the local view. In this work, a two-layer GCN is employed with the input feature matrix $\mathbf{X}$ and adjacency matrix $\mathbf{A}$, namely
\begin{equation}
\label{local_gcn}
\mathbf{H}^{\phi_1} = \hat{ \mathbf{A}}\sigma (\hat{\mathbf{A}}\mathbf{X}{\mathbf{W}^{(0)}}){\mathbf{W}^{(1)}},
\end{equation}
where $\hat{ \mathbf{A}} = {\tilde{\mathbf{D}}^{ - \frac{1}{2}}}\tilde{ \mathbf{A}}{\tilde{ \mathbf{D}}^{ - \frac{1}{2}}}$, $\tilde{ \mathbf{A}} = \mathbf{A} + \mathbf{I}$, ${\tilde{ \mathbf{D}}_{ii}} = \sum\nolimits_j {{{\tilde {\mathbf{A}}}_{ij}}}$, $\mathbf{W}^{(0)}$ and $\mathbf{W}^{(1)}$ denote the trainable weight matrices, $\sigma ( \cdot )$ represents an activation function (\emph{e.g.}, the ReLU function \cite{Nair2010Rectified}), and $\mathbf{H}^{\phi_1}$ denotes the representation result learned from view $\phi_1$ (\emph{i.e.}, the local view). 

Afterwards, we employ a simple yet effective hierarchical GCN model, i.e., HGCN \cite{hu2019hierarchical}, to generate the representations from the global view. Concretely, HGCN repeatedly aggregates the structurally similar graph nodes to a set of hyper-nodes, which can produce coarsened graphs for convolution and enlarge the receptive field for the nodes. Then, the symmetric graph refining layers are applied to restore the original graph structure for node-level representation learning. Such a hierarchical graph convolution model comprehensively captures the nodes' information from local to global perspectives. As a result, the representations $\mathbf{H}^{\phi_2}$ can be generated from the global view (\emph{i.e.}, view $\phi_2$), which provides complementary information to $\mathbf{H}^{\phi_1}$.

\subsection{Semi-Supervised Contrastive Learning}

Unsupervised contrastive methods have led to great success in various domains, as they can exploit rich information contained in the data themselves to guide the representation learning process. However, the unsupervised contrastive techniques \cite{hassani2020contrastive, zhu2020deep} fail to explore the class information which is scarce yet valuable in SSL problems. To address this issue, we propose a semi-supervised contrastive loss which incorporates the class information to improve the contrastive representation learning. The proposed semi-supervised contrastive loss can be partitioned into two parts, namely the supervised and unsupervised contrastive losses.

Formally, the unsupervised contrastive learning is expected to achieve the effect as
\begin{equation}
\label{contrastive}
{\rm{score}}(f({\mathbf{x}_i}),f(\mathbf{x}_i^ + )) \gg {\rm{score}}(f({\mathbf{x}_i}),f(\mathbf{x}_i^ - )),
\end{equation}
where $\mathbf{x}_i^ +$ is a node similar or congruent to $\mathbf{x}_i$, $\mathbf{x}_i^ -$ is a node dissimilar to $\mathbf{x}_i$, $f$ is an encoder, and the score function is used to measure the similarity of encoded features of two nodes. Here, ($\mathbf{x}_i$, $\mathbf{x}_i^ +$) and ($\mathbf{x}_i$, $\mathbf{x}_i^ -$) indicate the positive and negative pairs, respectively. Eq.~\eqref{contrastive} encourages the score function to assign large values to the positive pairs and small values to the negative pairs, which can be used as the supervision signals to guide the learning process of encoder $f$. By resorting to the above-mentioned explanations, our unsupervised contrastive loss ${\mathcal{L}_{uc}}$ can be presented as
\begin{equation}
\label{loss_uc}
{\mathcal{L}_{uc}} = \frac{1}{{2n}}\sum\limits_{i = 1}^n {(\mathcal{L}_{uc}^{\phi_1}({\mathbf{x}_i})  + \mathcal{L}_{uc}^{\phi_2}({\mathbf{x}_i}) )} ,
\end{equation}
where $\mathcal{L}_{uc}^{\phi_1}({\mathbf{x}_i})$ and $\mathcal{L}_{uc}^{\phi_2}({\mathbf{x}_i})$ denote the unsupervised pairwise contrastive losses of $\mathbf{x}_i$ in local and global views, respectively. Further, $\mathcal{L}_{uc}^{\phi_1}({\mathbf{x}_i})$ can be obtained with the similarity measured by inner product, namely
\begin{equation}
\label{L_v1_uc}
{\mathcal{L}_{uc}^{{\phi_1}}}({\mathbf{x}_i}) =  - \log \frac{{\exp (\langle\mathbf{h}_i^{{\phi_1}} , \mathbf{h}_i^{{\phi_2}}\rangle)}}{{\sum\nolimits_{j = 1}^n {\exp (\langle\mathbf{h}_i^{{\phi_1}} , \mathbf{h}_j^{{\phi_2}}\rangle)} }},
\end{equation}
where $\mathbf{h}_i^{\phi_1}=\mathbf{H}_{i,:}^{\phi_1}$ and $\mathbf{h}_i^{\phi_2}=\mathbf{H}_{i,:}^{\phi_2}$ denote the representation results of $\mathbf{x}_i$ learned from the local and global views, respectively, and $\langle\cdot\rangle$ denotes the inner product. Here $\mathbf{H}^{\phi_v}_{i,:}$ denotes the $i$-th row of the matrix $\mathbf{H}^{\phi_v}$ for $v=1,2$. By using Eq.~\eqref{L_v1_uc}, the similarity of the positive pairs (\emph{i.e.}, $\mathbf{h}_i^{{\phi_1}}$ and $\mathbf{h}_i^{{\phi_2}}$) can be contrasted with that of the negative pairs, and then $\mathcal{L}_{uc}^{\phi_2}({\mathbf{x}_i})$ can be similarly calculated by
\begin{equation}
\label{L_v1_uc_2}
{\mathcal{L}_{uc}^{{\phi_2}}}({\mathbf{x}_i}) =  - \log \frac{{\exp (\langle\mathbf{h}_i^{{\phi_2}} , \mathbf{h}_i^{{\phi_1}}\rangle)}}{{\sum\nolimits_{j = 1}^n {\exp (\langle\mathbf{h}_i^{{\phi_2}} , \mathbf{h}_j^{{\phi_1}}\rangle)} }}.
\end{equation}
To incorporate the scarce yet valuable class information for model training, we propose to use a supervised contrastive loss as follows:
\begin{equation}
\label{loss_sc}
{\mathcal{L}_{sc}} = \frac{1}{{2l}}\sum\limits_{i = 1}^l {(\mathcal{L}_{sc}^{{\phi_1}}({\mathbf{x}_i}) + \mathcal{L}_{sc}^{{\phi_2}}({\mathbf{x}_i}))}.
\end{equation}
Here, the supervised pairwise contrastive loss of $\mathbf{x}_i$ can be computed as
\begin{equation}
\label{sup_con_1}
\mathcal{L}_{sc}^{{\phi_1}}({\mathbf{x}_i}) =  - \log \frac{{\sum\nolimits_{k = 1}^l {{\mathbbm{1}_{[{y_i} = {y_k}]}}\exp (\langle\mathbf{h}_i^{{\phi_1}} , \mathbf{h}_k^{{\phi_2}}\rangle)} }}{{\sum\nolimits_{j = 1}^l {\exp (\langle\mathbf{h}_i^{{\phi_1}} , \mathbf{h}_j^{{\phi_2}}\rangle)} }},
\end{equation}
\begin{equation}
\label{sup_con_2}
\mathcal{L}_{sc}^{{\phi_2}}({\mathbf{x}_i}) =  - \log \frac{{\sum\nolimits_{k = 1}^l {{\mathbbm{1}_{[{y_i} = {y_k}]}}\exp (\langle\mathbf{h}_i^{{\phi_2}} , \mathbf{h}_k^{{\phi_1}}\rangle)} }}{{\sum\nolimits_{j = 1}^l {\exp (\langle\mathbf{h}_i^{{\phi_2}} , \mathbf{h}_j^{{\phi_1}}\rangle)} }},
\end{equation}
where ${\mathbbm{1}_{[\cdot]}}$ is an indicator function which equals to 1 if the argument inside the bracket holds, and 0 otherwise. Different from the unsupervised contrastive learning in Eqs.~\eqref{L_v1_uc} and \eqref{L_v1_uc_2}, here the positive and negative pairs are constructed based on the facts that whether two nodes belong to the same class. In other words, a data pair is positive if both examples have the same label, and is negative if their labels are different.

\begin{figure}[t]
	\centering
	\includegraphics[width=0.9\columnwidth]{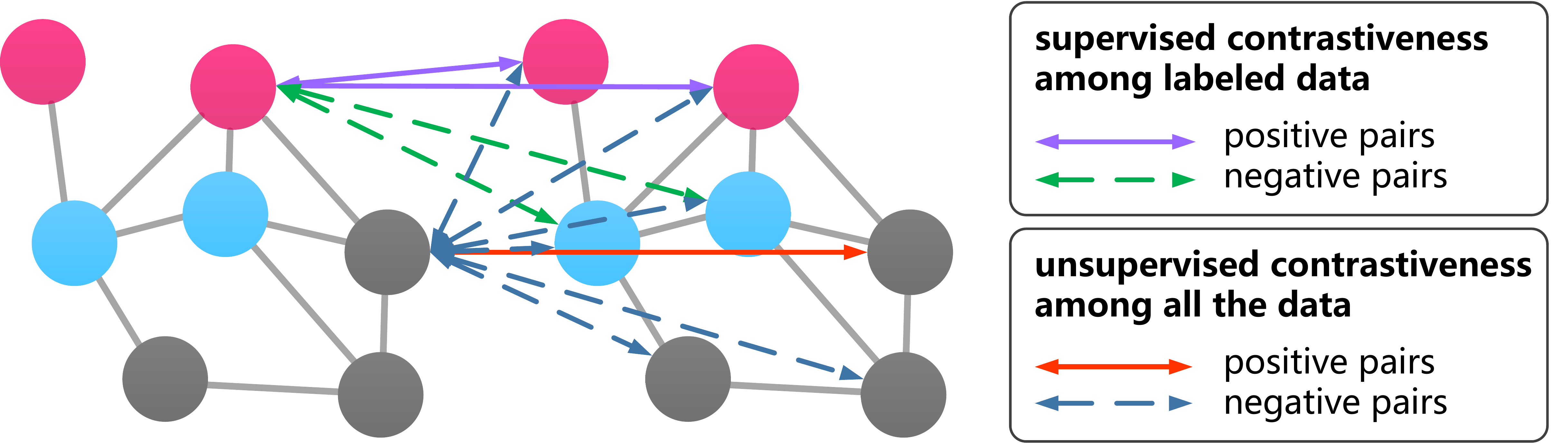} 
	\caption{The mechanism of our semi-supervised contrastive learning. The red and blue circles denote the labeled graph nodes, where each color corresponds to a specific class, and the gray circles represent the unlabeled graph nodes. Various contrastive strategies adopted by our method are illustrated by the arrows with different colors and line styles.}
	\label{links}
\end{figure}

By combining the supervised and unsupervised contrastive losses, we arrive at the following semi-supervised contrastive loss: 
\begin{equation}
\label{s2c}
{\mathcal{L}_{ssc}} = {\mathcal{L}_{uc}} + {\mathcal{L}_{sc}}.
\end{equation}
The mechanism of our semi-supervised contrastive learning has been exhibited in Figure~\ref{links}. By minimizing ${\mathcal{L}_{ssc}}$, the rich unlabeled data and the scarce yet valuable labeled data work collaboratively to provide additional supervision signals for discriminative representation learning, which can further improve the subsequent classification result.

\subsection{Graph Generative Loss}

Apart from the supervision information extracted from data similarities via contrastive learning, we also intend to distill the graph topological information to better guide the representation learning process. In this work, a graph generative loss is utilized to encode the graph structure and model the underlying relationship between the feature representations and graph topology. Inspired by the generative models \cite{hoff2002latent, kipf2016variational, ma2019flexible}, we let the graph edge $e_{ij}$ be the binary random variable with $e_{ij}=1$ indicating the existence of the edge between $\mathbf{x}_i$ and $\mathbf{x}_j$, and $e_{ij}=0$ otherwise. Here the edges are assumed to be conditionally independent, so that the conditional probability of the input graph $\mathcal{G}$ given $\mathbf{H}^{\phi_1}$ and $\mathbf{H}^{\phi_2}$ can be factorized as
\begin{equation}
\label{GenFunc_1}
{p }(\mathcal{G}|\mathbf{H}^{\phi_1}, \mathbf{H}^{\phi_2}) = \prod\limits_{i,j} {{p }({e_{ij}}|\mathbf{H}^{\phi_1}, \mathbf{H}^{\phi_2})}.
\end{equation}
Similar to the latent space models \cite{hoff2002latent, ma2019flexible}, we reasonably assume that the probability of $e_{ij}$ only depends on the representations of $\mathbf{x}_i$ and $\mathbf{x}_j$. Meanwhile, to further maximize the node-level agreement across the global and local views, the conditional probability of $e_{ij}$ can be obtained as ${{p }({e_{ij}}|\mathbf{H}^{\phi_1}, \mathbf{H}^{\phi_2})}={{p }({e_{ij}}|\mathbf{h}_i^{\phi_1}, \mathbf{h}_j^{\phi_2})}$. Finally, for practical use, we specify the parametric forms of the conditional probability by using a logit model, which arrives at
\begin{equation}
\label{instan}
 {p }(\mathcal{G}|\mathbf{H}^{\phi_1}, \mathbf{H}^{\phi_2}) =\prod\limits_{i,j} {{p }({e_{ij}}|\mathbf{h}_i^{\phi_1}, \mathbf{h}_j^{\phi_2})}=\prod\limits_{i,j}\delta  ([\mathbf{h}_i^{{\phi_1}},\mathbf{h}_j^{{\phi_2}}]\mathbf{w}),
\end{equation}
where $\delta(\cdot)$ is the logistic function, $\mathbf{w}$ is the learnable parameter vector, and $[\cdot,\cdot]$ is the concatenation operation. By maximizing Eq.~\eqref{instan}, the observed graph structure can be taken into consideration along with data feature and scarce label information for node classification, and thus the graph generative loss can be formulated as $\mathcal{L}_{g^2}=-{p }(\mathcal{G}|\mathbf{H}^{\phi_1}, \mathbf{H}^{\phi_2})$.

\subsection{Model Training}

To obtain the overall network output $\mathbf{O}$, we integrate the representation results generated by the GCN and HGCN models, so that the rich information from both local and global views can be exploited, which is expressed as
\begin{equation}
\mathbf{O} = \lambda^{\phi_1} \mathbf{H}^{\phi_1} + (1-\lambda^{\phi_1}) \mathbf{H}^{\phi_2}, 
\end{equation}
where $0<\lambda^{\phi_1}<1$ is the weight assigned to $\mathbf{H}^{\phi_1}$. Afterwards, the cross-entropy loss can be adopted to penalize the differences between the network output $\mathbf{O}$ and the labels of the originally labeled nodes as
\begin{equation}
\label{loss_cse}
{\mathcal{L}_{ce}} =  - \sum\limits_{i = 1}^l {\sum\limits_{j = 1}^c {{\mathbf{Y}_{ij}}\ln {\mathbf{O}_{ij}}} }.
\end{equation}
Finally, by combining $\mathcal{L}_{ce}$ with the semi-supervised contrastive loss $\mathcal{L}_{ssc}$ and the graph generative loss $\mathcal{L}_{g^2}$, the overall loss function of our CG$^3$ can be presented as
\begin{equation}
\label{loss_overall}
\mathcal{L} = {\mathcal{L}_{ce}} + \lambda_{ssc}{\mathcal{L}_{ssc}} + \lambda_{g^2}{\mathcal{L}_{g^2}},
\end{equation}
where $\lambda_{ssc}>0$ and $\lambda_{g^2}>0$ are tuning parameters to weight the importance of $\mathcal{L}_{ssc}$ and $\mathcal{L}_{g^2}$, respectively. The detailed description of our CG$^3$ is provided in Algorithm~\ref{Algorithm1}.

\begin{algorithm}[!t] 
	\caption{The Proposed CG$^3$ algorithm} 
	\label{Algorithm1} 
	\begin{algorithmic}[1] 
		\Require 
		Feature matrix $\mathbf{X}$; adjacency matrix $\mathbf{A}$; label matrix $\mathbf{Y}$; maximum number of iterations $\mathcal{T}$
		\For {$t=1$ to $\mathcal{T}$}
		\State // Multi-view representation learning
		\State Perform localized graph convolution (\emph{i.e.}, Eq.~\eqref{local_gcn}) and hierarchical graph convolution \cite{hu2019hierarchical} to obtain $\mathbf{H}^{\phi_1}$ and $\mathbf{H}^{\phi_2}$, respectively;
		\State // Calculate loss values
		\State Calculate semi-supervised contrastive loss ${\mathcal{L}_{ssc}}$ based on Eqs.~\eqref{loss_uc} and \eqref{loss_sc};
		\State Calculate the graph generative loss ${\mathcal{L}_{g^2}}$ based on Eq.~\eqref{instan};
		\State Calculate the cross-entropy loss ${\mathcal{L}_{ce}}$ with Eq.~\eqref{loss_cse};
		\State Update the network parameters according to the overall loss function $\mathcal{L}$ in Eq.~\eqref{loss_overall};
		\EndFor 
		\State \textbf{end for}
		\State Conduct label prediction based on the trained network;		
		\Ensure 
		Predicted label for each unlabeled graph node.
	\end{algorithmic} 
\end{algorithm}

\begin{table}[t]
	\centering
	\caption{Dataset statistics}
	\footnotesize
	\begin{tabular}{lrrrr}
		\toprule
		Datasets & Nodes & Edges & Features & Classes \\
		\midrule
		Cora  & 2,708 & 5,429 & 1,433 & 7 \\
		CiteSeer & 3,327 & 4,732 & 3,703 & 6 \\
		PubMed & 19,717 & 44,338 & 500   & 3 \\
		Amazon Computers & 13,752 & 245,861 & 767   & 10 \\
		Amazon Photo & 7,650 & 119,081 & 745   & 8 \\
		Coauthor CS & 18,333 & 81,894 & 6,805 & 15 \\
		\bottomrule
	\end{tabular}%
	\label{Datasets}%
\end{table}%

\section{Experimental Results}

To demonstrate the effectiveness of our proposed CG$^3$ method, extensive experiments have been conducted on six benchmark datasets including three widely-used citation networks (\emph{i.e.}, Cora, CiteSeer, and PubMed) \cite{sen2008collective, bojchevski2018deep}, two Amazon product co-purchase networks (\emph{i.e.}, Amazon Computers and Amazon Photo) \cite{shchur2018pitfalls}, and one co-author network subjected to computer science (\emph{i.e.}, Coauthor CS) \cite{shchur2018pitfalls}. Dataset statistics are summarized in Table \ref{Datasets}. We report the mean accuracy of ten independent runs for every algorithm on each dataset to achieve fair comparison.

\begin{table*}[t]
	\centering
	\caption{Classification accuracies of compared methods on Cora, CiteSeer, PubMed, Amazon Computers, Amazon Photo, and Coauthor CS datasets. Some records are not associated with standard deviations as they are directly taken from \cite{hassani2020contrastive} which did not report standard deviations.}
	\begin{tabular}{lcccccc}
		\toprule
		Method & Cora  & CiteSeer & PubMed & \makecell[c]{Amazon\\Computers} & \makecell[c]{Amazon\\Photo} & \makecell[c]{Coauthor\\CS} \\
		\midrule
		LP    & 68.0  & 45.3  & 63.0  & 70.8$\pm$0.0 & 67.8$\pm$0.0 & 74.3$\pm$0.0 \\
		Chebyshev & 81.2  & 69.8  & 74.4  & 62.6$\pm$0.0 & 74.3$\pm$0.0 & 91.5$\pm$0.0 \\
		GCN   & 81.5  & 70.3  & 79.0  & 76.3$\pm$0.5 & 87.3$\pm$1.0 & 91.8$\pm$0.1 \\
		GAT   & 83.0$\pm$0.7 & 72.5$\pm$0.7 & 79.0$\pm$0.3 & 79.3$\pm$1.1 & 86.2$\pm$1.5 & 90.5$\pm$0.7 \\
		SGC   & 81.0$\pm$0.0 & 71.9$\pm$0.1 & 78.9$\pm$0.0 & 74.4$\pm$0.1 & 86.4$\pm$0.0 & 91.0$\pm$0.0 \\
		DGI   & 81.7$\pm$0.6 & 71.5$\pm$0.7 & 77.3$\pm$0.6 & 75.9$\pm$0.6 & 83.1$\pm$0.5 & 90.0$\pm$0.3 \\
		GMI   & 82.7$\pm$0.2 & 73.0$\pm$0.3 & 80.1$\pm$0.2 & 76.8$\pm$0.1 & 85.1$\pm$0.1 & 91.0$\pm$0.0 \\
		MVGRL & 82.9$\pm$0.7 & 72.6$\pm$0.7 & 79.4$\pm$0.3 & 79.0$\pm$0.6 & 87.3$\pm$0.3 & 91.3$\pm$0.1 \\
		GRACE & 80.0$\pm$0.4 & 71.7$\pm$0.6 & 79.5$\pm$1.1 & 71.8$\pm$0.4 & 81.8$\pm$1.0 & 90.1$\pm$0.8 \\
		\midrule
		CG$^3$ & \textbf{83.4$\pm$0.7} & \textbf{73.6$\pm$0.8} & \textbf{80.2$\pm$0.8} & \textbf{79.9$\pm$0.6} & \textbf{89.4$\pm$0.5} & \textbf{92.3$\pm$0.2} \\
		\bottomrule
	\end{tabular}%
	\label{Table_all}%
\end{table*}%

\subsection{Node Classification Results}

We evaluate the performance of our proposed CG$^3$ method on transductive semi-supervised node classification tasks by comparing it with a variety of methods, including \textbf{L}abel \textbf{P}ropagation (LP) \cite{zhu2003semi}, Chebyshev \cite{defferrard2016convolutional}, GCN \cite{Kipf2016Semi}, GAT \cite{velivckovic2018graph}, SGC \cite{wu2019simplifying}, DGI \cite{velickovic2019deep}, GMI \cite{peng2020graph}, MVGRL \cite{hassani2020contrastive}, and GRACE \cite{zhu2020deep}. For the widely-used Cora, CiteSeer, and PubMed datasets, we use the same train/validation/test splits as \cite{yang2016revisiting}. For the other three datasets (\emph{i.e.}, Amazon Computers, Amazon Photo, and Coauthor CS), we use 30 labeled nodes per class as the training set, 30 nodes per class as the validation set, and the rest as the test set. Note that the selection of labeled nodes on each dataset is kept identical for all compared methods.

Classification results are reported in Table~\ref{Table_all}, where the highest record on each dataset has been highlighted in bold. Notably, the GCN-based contrastive models (\emph{i.e.}, DGI, GMI, MVGRL, GRACE, and CG$^3$) can generally achieve strong performance across all six datasets, which is due to the reason that contrastive learning aims to extract additional supervision information from data similarities for improving the learned representations, and thus obtaining promising classification results.  In our CG$^3$, two different types of GCNs are adopted to aggregate information from both local and global views. Meanwhile, CG$^3$ enriches the supervision signals from data similarities and graph structure simultaneously, which can help generate discriminative representations for classification tasks. Consequently, the proposed CG$^3$ consistently surpasses other contrastive methods and achieves the top level performance among all baselines on these six datasets.

\begin{table}[t]
	\centering
	\caption{Classification accuracies with different label rates on Cora dataset.}
	\begin{tabular}{lcccc}
		\toprule
		Label Rate & 0.5\% & 1\%   & 2\%   & 3\%   \\
		\midrule
		LP    & 56.4  & 62.3  & 65.4  & 67.5    \\
		Chebyshev & 36.4  & 54.7  & 55.5  & 67.3   \\
		GCN   & 42.6  & 56.9  & 67.8  & 74.9    \\
		GAT   & 56.4  & 71.7  & 73.5  & 78.5  \\
		SGC   & 43.7  & 64.3  & 68.9  & 71.0 \\
		DGI   & 67.5  & 72.4  & 75.6  & 78.9   \\
		GMI   & 67.1  & 71.0    & 76.1  & 78.8 \\
		MVGRL & 61.6  & 65.2  & 74.7  & 79.0 \\
		GRACE & 60.4  & 70.2  & 73.0  & 75.8  \\
		\midrule
		CG$^3$  & \textbf{69.3 } & \textbf{74.1 } & \textbf{76.6 } & \textbf{79.9 } \\
		\bottomrule
	\end{tabular}%

\label{Table_few_cora}%
\end{table}%

\subsection{Results under Scarce Labeled Training Data}

To further investigate the ability of our proposed CG$^3$ in dealing with scarce supervision, we conduct experiments when the number of labeled examples is extremely small. For each run, we follow \cite{li2018deeper} and select a small set of labeled examples for model training. The specific label rates are 0.5\%, 1\%, 2\%, 3\% for Cora and CiteSeer datasets, and 0.03\%, 0.05\%, 0.1\% for PubMed dataset. Here, the baselines are kept identical with the previous node classification experiments.

The results shown in Tables~\ref{Table_few_cora},~\ref{Table_few_citeseer}, and \ref{Table_few_pubmed} again verify the effectiveness of our CG$^3$ method. We see that CG$^3$ outperforms other state-of-the-art approaches under different small label rates across the three datasets. It can be observed that the performance of GCN significantly declines when the label information is very limited (\emph{e.g.}, at the label rate of 0.5\% on Cora dataset) due to the inefficient propagation of label information. In contrast, the GCN-based contrastive models (\emph{i.e.}, DGI, GMI, MVGRL, GRACE, and CG$^3$) can often achieve much better results with few labeled data, which demonstrates the benefits of extracting supervision information from data themselves to learn powerful representations for classification tasks. Besides, it is noteworthy that on each dataset, our CG$^3$ consistently outperforms the other GCN-based contrastive approaches (\emph{i.e.}, DGI, GMI, MVGRL, and GRACE) by a large margin, especially when the labeled data becomes very limited. This is due to that our proposed CG$^3$ can additionally exploit the supervision signals from graph topological and label information simultaneously, which has often been ignored by other contrastive models.

\begin{table}[t]
	\centering
	\caption{Classification accuracies with different label rates on CiteSeer dataset.}
	\begin{tabular}{lcccc}
		\toprule
		Label Rate & 0.5\% & 1\%   & 2\%   & 3\% \\
		\midrule
		LP    & 34.8  & 40.2  & 43.6  & 45.3  \\
		Chebyshev & 19.7  & 59.3  & 62.1  & 66.8  \\
		GCN   & 33.4  & 46.5  & 62.6  & 66.9  \\
		GAT   & 45.7  & 64.7  & 69.0  & 69.3  \\
		SGC   & 43.2  & 50.7  & 55.8  & 60.9  \\
		DGI   & 60.7  & 66.9  & 68.1  & 69.8  \\
		GMI   & 56.2  & 63.5  & 65.7  & 68.0  \\
		MVGRL & 61.7  & 66.6  & 68.5  & 70.3  \\
		GRACE & 55.4  & 59.3  & 63.4  & 67.8  \\
		\midrule
		CG$^3$  & \textbf{62.7 } & \textbf{70.6 } & \textbf{70.9 } & \textbf{71.3 } \\
		\bottomrule
\end{tabular}%
\label{Table_few_citeseer}%
\end{table}%

\begin{table}[t]
	\centering
	\caption{Classification accuracies with different label rates on PubMed dataset.}
	\begin{tabular}{lccc}
		\toprule
		Label Rate & 0.03\% & 0.05\% & 0.1\% \\
		\midrule
		LP    & 61.4  & 65.4  & 66.4  \\
		Chebyshev & 55.9  & 62.5  & 69.5  \\
		GCN   & 61.8  & 68.8  & 71.9  \\
		GAT   & 65.7  & 69.9  & 72.4  \\
		SGC   & 62.5  & 69.4  & 69.9  \\
		DGI   & 60.2  & 68.4  & 70.7  \\
		GMI   & 60.1  & 62.4  & 71.4  \\
		MVGRL & 63.3  & 69.4  & 72.2  \\
		GRACE & 64.4  & 67.5  & 72.3  \\
		\midrule
		CG$^3$  & \textbf{68.3 } & \textbf{70.1 } & \textbf{73.2 } \\
		\bottomrule
	\end{tabular}%
\label{Table_few_pubmed}%
\end{table}%

\begin{table}[t]
	\centering	
	\caption{Ablation study of the contrastive and generative losses on Cora, CiteSeer, and PubMed datasets.}
	\begin{tabular}{lccc}
		\toprule
		Method & Cora  & CiteSeer & PubMed \\
		\midrule
		CG$^3$ (w/o ConLoss) & 79.2$\pm$0.7 & 69.8$\pm$1.3 & 76.6$\pm$1.0 \\
		CG$^3$ (w/o GenLoss) & 82.9$\pm$0.9 & 72.9$\pm$0.9 & 79.8$\pm$0.9 \\
		\midrule
		CG$^3$ & \textbf{83.4$\pm$0.7} & \textbf{73.6$\pm$0.8} & \textbf{80.2$\pm$0.8} \\
		\bottomrule
	\end{tabular}%
	
	\label{Table_ablation}%
\end{table}%

\begin{figure}[t]
	\centering
	\subfigure[]{%
		\label{tsne_cg3}
		\resizebox*{2.3cm}{!}{\includegraphics{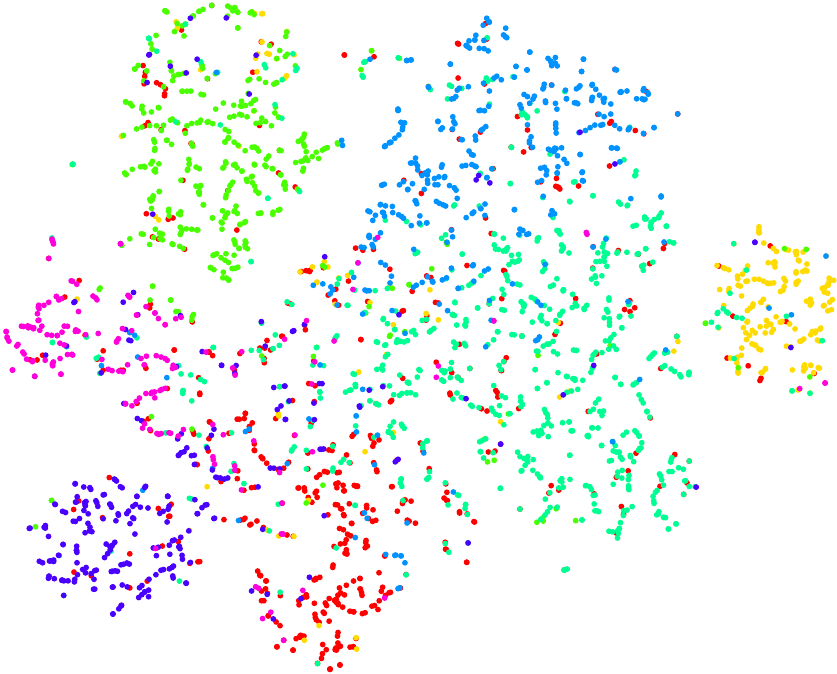}}}\hspace{5pt}
	\subfigure[]{%
		\label{tsne_gcn}
		\resizebox*{2.3cm}{!}{\includegraphics{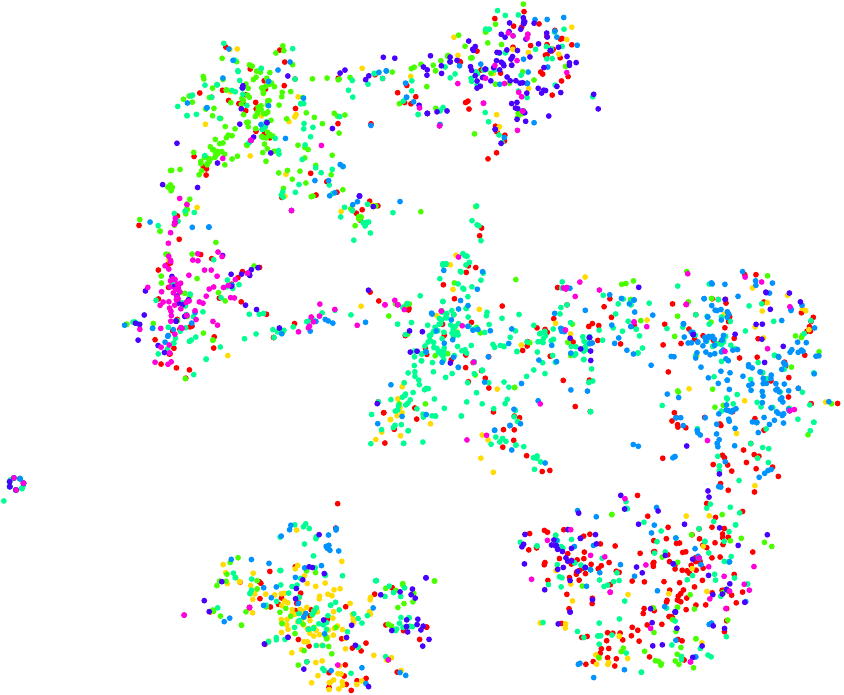}}}\hspace{5pt}
	\subfigure[]{%
		\label{tsne_hgcn}
		\resizebox*{2.3cm}{!}{\includegraphics{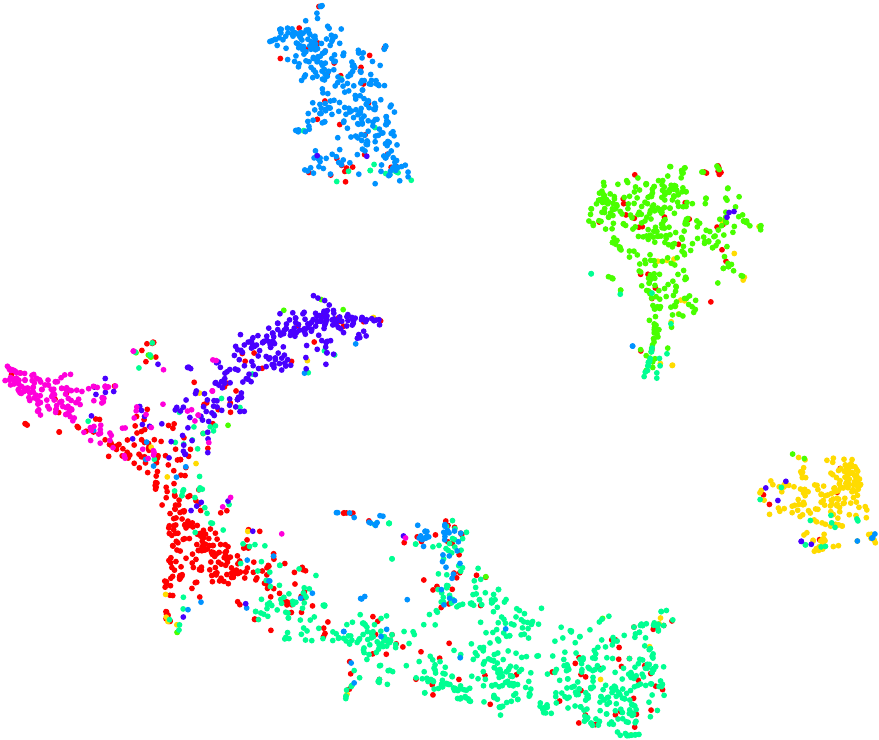}}}\hspace{0pt}
	\caption{t-SNE embeddings of the nodes acquired by different methods on Cora dataset. (a) GCN; (b) HGCN; (c) CG$^3$.}
	\label{ablation_tsne}
\end{figure}

\subsection{Ablation Study}

As is mentioned in the introduction, our proposed CG$^3$ employs the contrastive and graph generative losses to enrich the supervision signals from the data similarities and graph structure, respectively. To shed light on the contributions of these two components, we report the classification results of CG$^3$ when each of the two components is removed on the three previously-used datasets including Cora, CiteSeer, and PubMed. The data splits are kept identical with \cite{yang2016revisiting}. For simplicity, we adopt `CG$^3$ (w/o ConLoss)' and `CG$^3$ (w/o GenLoss)' to represent the reduced models by removing the contrastive loss $\mathcal{L}_{ssc}$ and the graph generative loss $\mathcal{L}_{g^2}$, respectively, and the comparative results have been exhibited in Table~\ref{Table_ablation}. It is apparent that the classification accuracy will decrease when any one of the aforementioned components is dropped, which reveals that both components make essential contributions to boosting the performance. In particular, our proposed model is able to improve the classification performance substantially by utilizing the contrastive loss, \emph{e.g.}, the accuracy can be raised by nearly 4\% on CiteSeer dataset.

Meanwhile, it is noteworthy that our proposed model performs graph convolution in different views based on two parallel networks (\emph{i.e.}, GCN and HGCN), and also conducts contrastive operation between these two views. As a result, the abundant local and global information are encoded simultaneously to obtain the improved data representations for classification. To reveal this, we visualize the embedding results of Cora dataset generated by GCN, HGCN, and CG$^3$ via using t-SNE method \cite{maaten2008visualizing}, which are given in Figure~\ref{ablation_tsne}. As can be observed, the 2D projections of the embeddings generated by our CG$^3$ (see Figure~\ref{tsne_hgcn}) can exhibit more coherent clusters when compared with the other two methods. Therefore, we believe that the contrastiveness among multi-view graph convolutions is beneficial to rendering promising classification results.

\section{Conclusion}

In this paper, we have presented the \textbf{C}ontrastive \textbf{G}CNs with \textbf{G}raph \textbf{G}eneration (CG$^3$) which is a new GCN-based approach for transductive semi-supervised node classification. By designing a semi-supervised contrastive loss, the scarce yet valuable class information, together with the data similarities, can be used to provide abundant supervision information for discriminative representation learning. Moreover, the supervision signals can be further enriched by leveraging the underlying relationship between the input graph topology and data features. Experiments on various public datasets illustrate the effectiveness of our method in solving different kinds of node classification tasks.

\bibliography{aaai21}
\bibstyle{aaai21}
\end{document}